\documentclass[a4paper,10pt]{article}

\usepackage{amsmath,amssymb,natbib,graphicx,sectsty,color,tikz}

\usetikzlibrary{arrows}

\renewcommand{\vec}{\boldsymbol}
\newcommand{\N}{\mathcal{N}}
\newcommand{\Trans}{^{\text{T}}}
\renewcommand{\d}{\operatorname{d}\!}

\newcommand{\cov}{\operatorname{cov}}
\newcommand{\wo}{\setminus}
\newcommand{\Ind}{\mathbb{I}}

\newcommand{\gen}{\mathfrak{g}}
\newcommand{\mi}{\mathfrak{m}}
\newcommand{\com}{\mathfrak{c}}
\newcommand{\I}{\mathfrak{I}}

\allsectionsfont{\sffamily}
\makeatletter  
\long\def\@makecaption#1#2{%
  \vskip\abovecaptionskip
  \sbox\@tempboxa{{\captionfonts #1: #2}}%
  \ifdim \wd\@tempboxa >\hsize
    {\captionfonts #1: #2\par}
  \else
    \hbox to\hsize{\hfil\box\@tempboxa\hfil}%
  \fi
  \vskip\belowcaptionskip}
\makeatother   

\newcommand{\captionfonts}{\sf }

\tikzset{>=stealth'}
\tikzstyle{var}   =[circle,draw=black,fill=white,minimum size=20pt]
\tikzstyle{obs}   =[circle, draw=black,fill=black,minimum size=20pt,text=white]
\tikzstyle{factor}=[rectangle,draw=black,fill=black!25,minimum size=20pt]
\tikzstyle{edge}  =[draw,-]
\tikzstyle{pre}   =[draw,->]
\tikzstyle{post}  =[draw,<-]
\tikzstyle{prior} =[rectangle, draw=black, fill=black, minimum size = 5pt]

\begin{document}

\title{\sffamily \bfseries Expectation Propagation on the Maximum\\ of Correlated Normal Variables}

\author{\sf Philipp Hennig\\ \sf Cavendish Laboratory\\\sf University of Cambridge\\
\sf CB3 0HE Cambridge, UK \\{\tt ph347@cam.ac.uk}}

\date{\sf July 2009}
\maketitle

\begin{abstract}
  Many inference problems involving questions of optimality ask for
  the maximum or the minimum of a finite set of unknown quantities. This technical report
  derives the first two posterior moments of the maximum 
  of two correlated Gaussian variables and the first two posterior moments of the two generating
  variables (corresponding to Gaussian approximations minimizing
  relative entropy).  It is shown how this can be used to build a heuristic
  approximation to the maximum relationship over a finite set of
  Gaussian variables, allowing approximate inference by Expectation Propagation on
  such quantities.
\end{abstract}

\section{Introduction}
\label{sec:Intro}

Many optimization problems involve inference on the maximum or minimum
of a set of variables. This very broad class includes shortest path
problems \citep{burton1992iis}, Reinforcement Learning \citep{BayesQ},
and scientific inference in Seismology \citep{neumanndenzau1984isd},
to name but a few. Often, there is a corresponding inverse
optimization problem \citep{ahuja2001io,heuberger2004ico}, where the
optimal solution is known with some uncertainty and the question is
about the quantities generating this optimum. Most contemporary
algorithms for this case aim to provide a point estimate (typically
the least-squares solution), but have trouble offering an error
estimate on this estimate as well.

This work derives (Section \ref{sec:max_two}) mean and variance 
of the posterior of the maximum of two correlated
Gaussian variables (for forward optimization problems), and the mean
and variance on the posterior of the Gaussian variables generating the
maximum (for inverse optimization problems). These two moments
correspond to the approximation within the exponential family of
Gaussian distributions minimizing the Kullback-Leibler Divergence
(relative entropy) to the true posterior. It will be shown how these 
results can be used to build a heuristic approximation to the max of a finite set of
normal variables (Section \ref{sec:max_many}). Together, this provides
the necessary results for Expectation Propagation \citep{minka2001epa}
on graphs involving the ``max'' relationship. Because maximum and
minimum obey the simple relationship $\max(\{x_i\}) = -
\min(\{-x_i\})$, this also allows inference on the minimum where
necessary. Limitations of this approximation are examined in Section
{\ref{sec:Discussion}.

  The moments of the normalized likelihood function of the maximum of
  two normal variables have previously been derived by
  \citet{ClarkMax}. To my best knowledge, this is the first
  publication deriving the full posterior, and the first to report the
  posterior for the inverse problem (see also Section \ref{sec:related_work}).

\section{The Maximum of Two Gaussian Variables}
\label{sec:max_two}

\subsection{Notation}
We consider two normally distributed variables $x_1$ and $x_2$,
forming the vector $\vec{x}$. Let there be some prior (i.e. outside)
information $\I_\gen$ giving rise to the belief
\marginpar{$\vec{x},\I_\gen$}
\begin{equation}
  \label{eq:1}
  \begin{aligned}
    p(x_{1},x_{2}|\I_\gen) &= \N(\vec{x};\vec{\mu}_\gen,\vec{\Sigma}_\gen) = \frac{1}{2\pi \sigma_{\gen1}\sigma_{\gen2} (1-\rho^{2})^{1/2}} \exp \left( -\frac{1}{2} (\vec{x}-\vec{\mu}_\gen)\Trans\vec{\Sigma}_\gen ^{-1}(\vec{x}-\vec{\mu}_\gen) \right)
  \end{aligned}
\end{equation}
over their values. Here we have defined a mean vector
$\vec{\mu}_\gen=(\mu_{\gen1},\mu_{\gen2})\Trans$ and a covariance
matrix $\vec{\Sigma}_\gen$. The latter has the form
\begin{equation}
  \label{eq:2}
  \vec{\Sigma}_\gen =
  \begin{pmatrix}
    \sigma_{\gen1} ^{2} & \rho \sigma_{\gen1} \sigma_{\gen2} \\ \rho
    \sigma_{\gen1} \sigma_{\gen2} & \sigma_{\gen2} ^{2}
  \end{pmatrix} \qquad \text{and thus} \qquad \vec{\Sigma}_\gen ^{-1}
  = \frac{1}{\sigma_{\gen1}^{2}\sigma_{\gen2}^{2}(1 -
    \rho^{2})} \begin{pmatrix} \sigma_{\gen2} ^{2} & -\rho
    \sigma_{\gen1} \sigma_{\gen2} \\ -\rho \sigma_{\gen1}
    \sigma_{\gen2} & \sigma_{\gen1} ^{2}
  \end{pmatrix}
\end{equation}
with the \emph{linear coefficient of correlation}\marginpar{$\rho$}
\begin{equation}
  \label{eq:3}
  \rho = \frac{\cov(x_{1},x_{2})}{\sigma_{\gen1}\sigma_{\gen2}}
\end{equation}
(for notational convenience, the index $\gen$ is dropped from $\rho$
because there will be no chance for confusion). We further introduce
the variable $m$ which is defined through $m=\max(x_1,x_2)$, and we
assume that there is some outside prior information $\I_\mi$ on the
value of $m$ as well: \marginpar{$m,\I_\mi$}
\begin{equation}
p(m|\I_\mi) = \N(m;\mu_\mi,\sigma_\mi ^2)
\label{eq:4}
\end{equation}
The inference problems to be solved are 
\begin{itemize}
	\item The posterior over $m$ given both $\I_\mi$ and $\I_\gen$ (jointly called $\I_\com$):
	\begin{equation}
			p(m|\I_\com) = \frac{p(m|\I_\mi)\int p(\vec{x}|m)p(\vec{x}|\I_\gen)\d \vec{x}}{\int \left[ p(m|\I_\mi)\int p(\vec{x}|m)p(\vec{x}|\I_\gen)\d \vec{x} \right] \d m}
			= Z^{-1} p(m|\I_\mi)\int p(\vec{x}|m)p(\vec{x}|\I_\gen)\d \vec{x}
		\label{eq:5}
	\end{equation}
	with the normalization constant $Z=\iint p(\vec{x},m|\I_\com) \d \vec{x} \d m$. This problem will be called the ``forward'' problem here.
	\item The posterior over $\vec{x}$ given $\I_\com$, 
		\begin{equation}
			p(\vec{x}|\I_\com) = \frac{p(\vec{x}|\I_\gen)\int p(m|\vec{x}) p(m|\I_\mi) \d m }{\int \left[ p(\vec{x}|\I_\gen)\int p(\vec{x}|m) p(m|\I_\mi) \d m \right]\d \vec{x}}
			= Z^{-1} p(\vec{x}|\I_\gen)\int p(m|\vec{x}) p(m|\I_\mi) \d m 
		\label{eq:6}
	\end{equation}
	This problem will be called the ``inverse'' problem.
\end{itemize}
Throughout the derivations\marginpar{$\N,\phi,\Phi$}, the notation 
\begin{equation}
  \label{eq:7}
  \begin{aligned}
    \N(x;\mu,\sigma^{2}) &\equiv \frac{1}{\sqrt{2\pi\sigma^{2}}} \exp\left[-\frac{1}{2} \left(\frac{x-\mu}{\sigma}\right)^{2} \right]\\
    \phi(x) &\equiv \frac{1}{\sqrt{2\pi}} \exp\left(-\frac{x^{2}}{2} \right)\\
    \Phi(x) &\equiv \int_{-\infty} ^{x} \phi(t)\d t = \frac{1}{2}
    \left[1 + \operatorname{erf}\left(\frac{x}{\sqrt{2}}\right)\right]
\end{aligned}
\end{equation}
will be used to denote the general and standard normal probability
density functions (PDF) and the standard normal cumulative distribution
function (CDF).

\subsection{Some Integrals}
The derivations in this paper will repeatedly feature certain
integrals. The first two incomplete moments of the standard Gaussian
are
\begin{equation}
  \label{eq:im_std}
  \begin{aligned}[t]
    \int_{-\infty} ^{y} t \phi(t)\d t &= -\phi(y)\\
    \int_{-\infty} ^{y} t^{2} \phi(t)\d t &= \Phi(y) - y \phi(y)
  \end{aligned}
\end{equation}
this is obvious directly from differentiation. A simple substitution
gives
\begin{alignat}{6}
  \int_{-\infty} ^y t \N(t;\alpha,\beta^2) \d t&= &\alpha &\Phi\left(
    \frac{y-\alpha}{\beta} \right) - & \beta &\phi\left(\frac{y-\alpha}{\beta} \right)\\
  \int_{-\infty} ^y t^2 \N(t;\alpha,\beta^2) \d t&= &(\alpha^2 +
  \beta^2)&\Phi\left(\frac{y-\alpha}{\beta}\right) -& (\alpha +
  y)\beta &\phi\left(\frac{y-\alpha}{\beta}\right)
\label{eq:im_gen}
\end{alignat}
Further, we will use the integrals
\begin{equation}
\begin{aligned}
  \int _{-\infty} ^\infty \Phi\left(\frac{x-a}{b}\right) \N(x;\alpha,\beta^2) \d x &= \Phi(z)\\
  \int _{-\infty} ^\infty x\Phi\left(\frac{x-a}{b}\right) \N(x;\alpha,\beta^2) \d x &= \alpha \Phi(z) + \frac{\beta^2}{b\sqrt{1 + \beta^2/b^2}} \phi(z)\\
  \int _{-\infty} ^\infty x^2 \Phi\left(\frac{x-a}{b}\right)
  \N(x;\alpha,\beta^2) \d x &= (\alpha^2 + \beta^2) \Phi(z) +
  \left[2\alpha \frac{\beta^2}{b\sqrt{1 + \beta^2/b^2}} - z \frac{\beta^4}{b^2 + \beta^2}\right] \phi(z) \\
  \text{where} \qquad z &= \frac{\alpha-a}{b\sqrt{1 + \beta^2/b^2}}
\end{aligned}
\label{eq:tm_gen}
\end{equation}
A derivation of these results can, for example, be found in
\citet[section 3.9]{RasmussenWilliams}

\subsection{Analytic Forms}
\label{sec:Analytic}

\subsubsection{Forward Problem}
Neither of the posterior distributions are normal themselves. The
forward posterior is\marginpar{$\nu_1,\nu_2$}
\begin{equation}
  \label{eq:8}
  \begin{aligned}
    p(m|\I_\com) &= Z ^{-1} p(m|\I_\mi)\iint_{-\infty} ^{\infty} p(\vec{x}|m) p(\vec{x}|\I_\gen) \d \vec{x}\\
    &= Z ^{-1} p(m|\I_\mi)\int_{-\infty}
    ^{\infty}\left[\int_{-\infty} ^{x_{1}} \delta(x_{1}-m)
      p(\vec{x}|\I_\gen) \d x_{2} +\int_{x_{1}} ^{\infty} \delta(x_{2}
      - m)
      p(\vec{x}|\I_\gen) \d x_{2} \right] \d x_{1} \\
    &= Z ^{-1} \mathop{\underbrace{p(m|\I_\mi)\int_{-\infty}
        ^{\infty} \delta(x_{1}-m) \int_{-\infty} ^{x_{1}}
        p(\vec{x}|\I_\gen) \d x_{2} \d x_{1} }}_{\nu_{1}} \\
        &\quad+ Z ^{-1}
    \mathop{\underbrace{p(m|\I_\mi)\int_{-\infty} ^{\infty}
        \delta(x_{2} -m) \int_{-\infty} ^{x_{2}} p(\vec{x}|\I_\gen) \d x_{1} \d x_{2}}}_{\nu_{2}} \\
  \end{aligned}
\end{equation}
For a motivation of the change in the integration ranges from the
second to the third line in Equation \eqref{eq:8}, consider the sketch
in Figure \ref{fig:integration}.
\begin{figure}[ht]
	\begin{center}
    \begin{tikzpicture}[thick]
     \fill[fill=black!20] (-2,2) -- (2,2) -- (-2,-2);
     \draw[->] (-2,0) -- (2,0);
     \node at (2,0.2) {$x_1$};
     \draw[->] (0,-2) -- (0,2);
     \node at (0.3,1.8) {$x_2$};
  	\end{tikzpicture}
	\end{center}
	\caption{Sketch of the integration range for $\nu_2$. The
          open set $(x_1,x_2)\in ((-\infty,\infty),(x_1,\infty))$ is
          identical to the open set $(x_1,x_2)\in
          ((-\infty,x_2),(-\infty,\infty))$.}
	\label{fig:integration}
\end{figure}
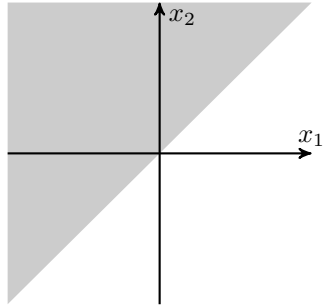
Since the two summands are related to each other through the symmetry
$x_{1} \leftrightarrow x_{2}$, consider only the first term,
$\nu_{1}$. To solve the integrals, note that the bi-variate Gaussian
$p(\vec{x}|\I_\gen)$ can be re-written as
\begin{equation}
  \label{eq:9}
  \begin{aligned}
  p(x_{1},x_{2}|\I_\gen) &= p(x_{1}|\I_\gen) p(x_{2}|x_1,\I_\gen) \\
  &= \frac{1}{\sqrt{2\pi \sigma_{\gen1}^{2}}} \exp\left[-\frac{1}{2} \left( \frac{x_{1}-\mu_{\gen1}} {\sigma_{\gen1}}\right)^{2} \right] \\
  &\quad  \frac{1} {\sqrt{2\pi \sigma_{\gen2}^{2} (1-\rho^{2})}} \exp \left[ -\frac{1}{2\sigma_{\gen2}^2(1-\rho^{2})} \left(x_{2}-\left(\mu_{\gen 2} + \rho\frac{\sigma_{\gen 2}}{\sigma_{\gen 1}}(x_{1}-\mu_{\gen 1})\right)\right)^{2} \right]
  \end{aligned}
\end{equation}
So we can simplify $\nu_{1}$ to 
\begin{equation}
  \label{eq:10}
\begin{aligned}
  \nu_{1} &= p(m|\I_\mi)\N(m;\mu_{\gen1},\sigma^{2} _{\gen1})
  \int_{-\infty} ^{m} \frac{1} {\sqrt{2\pi \sigma_{\gen2}^{2}
      (1-\rho^{2})}} \exp \left[ -\frac{1}{2(1-\rho^{2})}
    \left(\frac{x_{2}-\mu_{\gen2}}{\sigma_{\gen2}} -
      \rho\frac{m-\mu_{\gen1}}{\sigma_{\gen1}}\right)^{2} \right] \d x_{2} \\
  &= p(m|\I_\mi)\N(m;\mu_{\gen1},\sigma^{2} _{\gen1}) \int_{-\infty}
  ^{m} \frac{1} {\sqrt{2\pi \sigma_{\gen2}^{2} (1-\rho^{2})}} \exp
  \left[ -\frac{1}{2} \left(\frac{x_{2} - \mu_{\gen2} -
        \rho\frac{\sigma_{\gen2}}{\sigma_{\gen1}}
        (m-\mu_{\gen1})}{\sigma_{\gen2}(1-\rho^{2})^{1/2}}\right)^2
  \right] \d x_{2}
\end{aligned}
\end{equation}
We introduce the substitution
\begin{equation}
  \label{eq:11}
  t(x_{2}) \equiv \frac{x_{2} - \mu_{\gen2} - \rho\frac{\sigma_{\gen2}}{\sigma_{\gen1}} (m-\mu_{\gen1})}{\sigma_{\gen2}(1-\rho^{2})^{1/2}} 
  \quad\text{with Jacobian}\quad \frac{\d t}{\d x_2} = \frac{1}{\sigma_{2}(1-\rho^{2})^{1/2}}
\end{equation}
Which allows us to solve the integral and find the posterior up to
normalization
\begin{equation}
  \label{eq:12}	
  	\begin{aligned}
          p(m|\I_\com)&= Z^{-1} \N(\mu_\mi;\mu_{\gen 1},\sigma_\mi ^2 + \sigma^2 _{\gen 1})\N(m;\mu_{\com 1},\sigma^{2} _{\com 1}) \Phi \left( \frac{(\sigma_{\gen 1} - \rho \sigma_{\gen 2})m - \sigma_{\gen 1} \mu_{\gen 2} + \rho \sigma_{\gen 2} \mu_{\gen 1}}{\sigma_{\gen 1}\sigma_{\gen 2} (1-\rho^{2})^{1/2}} \right) \\
          &\quad + Z ^{-1} \N(\mu_\mi;\mu_{\gen 2},\sigma_\mi ^2 +
          \sigma^2 _{\gen 2}) \N(m;\mu_{\com 2},\sigma^{2} _{\com 2})
          \Phi \left( \frac{(\sigma_{\gen 2} - \rho \sigma_{\gen 1})m
              - \sigma_{\gen 2} \mu_{\gen 1} + \rho \sigma_{\gen 1}
              \mu_{\gen 2}}{\sigma_{\gen 2}\sigma_{\gen 1}
              (1-\rho^{2})^{1/2}} \right)
  \end{aligned}
\end{equation}
Where we have used \marginpar{$\mu_{\com},\sigma_{\com} ^2$} the abbreviations
\begin{equation}
  \sigma^2 _{\com 1} \equiv \frac{\sigma^2 _{\gen 1}  \sigma^2 _{\mi}}{\sigma^2 _{\com 1} + \sigma^2 _\mi } \qquad \text{and}\qquad  \mu_{\com 1} \equiv \left(\frac{\mu_{\gen 1}}{\sigma^2 _{\gen 1}} + \frac{\mu_\mi}{\sigma^2 _\mi} \right) \sigma^2 _{\com 1}
\label{eq:13}
\end{equation}
for the mean and variance of the product of two Gaussians\footnote{This is using
the standard result that
\begin{equation}
  \N(x;a_1,b^2 _1)\N(x;a_2,b^2 _2) = \N(a_1;a_2,b_1 ^2 + b_2 ^2) \N\left[x; \left(\frac{a_1}{b_1 ^2} + \frac{a_2}{b_2 ^2 }\right)\left(\frac{1}{b_1 ^2} + \frac{1}{b_2 ^2}\right)^{-1}, \left(\frac{1}{b_1 ^2} + \frac{1}{b_2 ^2}\right)^{-1} \right]
\label{eq:g_prod}
\end{equation}
which can be derived by completing the square, a simple proof that is
omitted here}, and
analogously for $\mu_{\com 2}$ and $\sigma_{\com 2}$. To find the normalization constant $Z$, we use the
first identity in Equation \eqref{eq:tm_gen} to get \marginpar{Z}
\begin{equation}
  Z = \N(\mu_{\mi};\mu_{\gen 1},\sigma_{\mi} ^{2} + \sigma_{\gen 1} ^{2}) \Phi(k_{1}) + \N(\mu_{\mi};\mu_{\gen 2},\sigma_{\mi} ^{2} + \sigma_{\gen 2} ^{2}) \Phi(k_{2})
\label{eq:m_Z}
\end{equation}
with \marginpar{$k_1,k_2$}
\begin{equation}
  k_{1} = \frac{(\sigma_{\gen 1}-\rho\sigma_{\gen 2})\mu_{\com 1} -
    \sigma_{\gen 1}\mu_{\gen 2} + \rho \sigma_{\gen 2}\mu_{\gen 1}}{\left[
      \sigma_{\gen 1}^{2}\sigma_{\gen 2}^{2}(1-\rho^{2}) +
      (\sigma_{\gen 1}-\rho\sigma_{\gen 2})^{2}\sigma_{\com 1} ^{2}
    \right]^{1/2}} 
  \quad\text{and}\quad 
  k_{2} =\frac{(\sigma_{\gen 2} - \rho \sigma_{\gen
      1})\mu_{\com 2}  -\sigma_{\gen 2}  \mu_{\gen 1} + \rho \sigma_{\gen 1}
    \mu_{\gen 2}}{\left[ \sigma_{\gen 1} ^2 \sigma_{\gen
        2} ^2 (1-\rho^2) + (\sigma_{\gen 2} - \rho
      \sigma_{\gen 1})^2  \sigma^2 _{\com 2} \right]^{1/2}}
\label{eq:k_norm}
\end{equation}

\subsubsection{Inverse Problem}
The conditional probability of $\vec{x}$ on $m$ is 
\begin{equation}
  p(x_1,x_2|m) = \Theta(x_1-x_2) \delta(x_1 - m) + \Theta(x_2-x_1) \delta(x_2 - m) 
\label{eq:14}
\end{equation}
where $\Theta(y)$ is Heaviside's step function. Therefore, the conditional of $\vec{x}$ on $\I_\mi$
is
\begin{equation}
\begin{aligned}
  p(x_1,x_2|\I_\mi) &= \int_{-\infty} ^\infty p(m|x_1,x_2) p(m|\I_\mi) \d m\\
  &= \Theta(x_1-x_2) \N(x_1;\mu_\mi,\sigma_\mi ^2) + \Theta(x_2 - x_1)
  \N(x_2;\mu_\mi,\sigma_\mi ^2)
 \end{aligned}
\label{eq:15}
\end{equation}
which is a proper (i.e. normalizable) distribution, but becomes normalizable after multiplication with the prior:
\begin{equation}
\begin{aligned}
  p(\vec{x}|\I_\com) &= Z ^{-1} \mathop{\underbrace{\Theta(x_1-x_2) \N(x_1;\mu_\mi,\sigma_\mi ^2) \N(\vec{x};\vec{\mu}_\gen,\vec{\Sigma} _\gen)}}_{\xi_1} \\
  &\quad+ Z ^{-1} \mathop{\underbrace{\Theta(x_2 - x_1) \N(x_2;\mu_\mi,\sigma_\mi ^2) \N(\vec{x};\vec{\mu}_\gen,\vec{\Sigma}_\gen)}}_{\xi_2}
  \end{aligned}
\label{eq:16}
\end{equation}
Figure \ref{fig:illustrations} illustrates the shape of these functions by way of some concrete examples.
\begin{figure}[ht]
  \centering
   \includegraphics[width=0.49\textwidth]{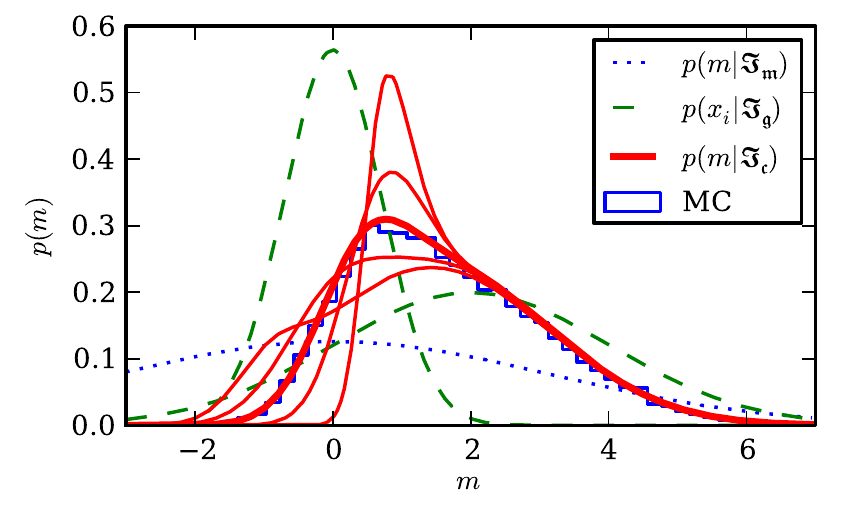}
  \includegraphics[width=0.49\textwidth]{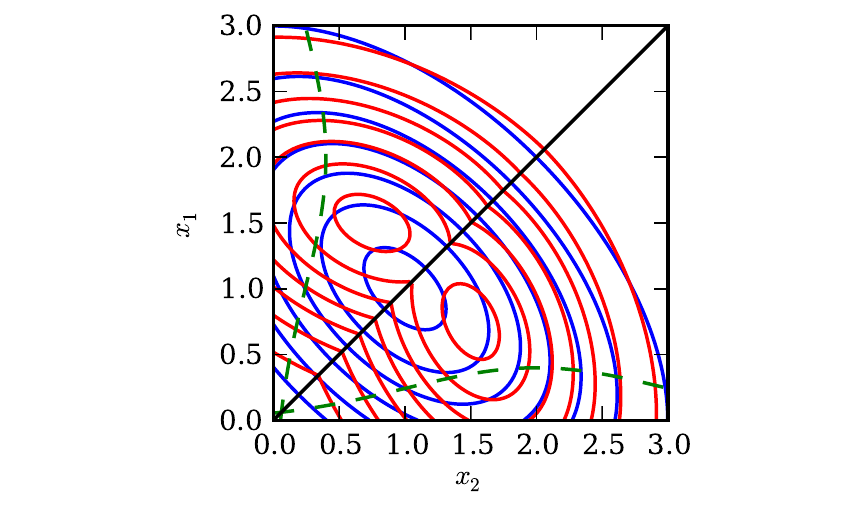}
  \caption{Illustrative plots for the analytical form of the forward
    and inverse posteriors. Left: Inference on $m$. Prior distribution
    and marginals on $\vec{x_{i}}$. Posteriors for five different values of $\rho$: -0.9
    (most peaked), -0.5, 0.0 (thick line), 0.5 and 0.9 (broadest). As
    an experimental verification, a histogram of 20,000 samples from
    the posterior (generated by rejection sampling, with $\rho=0$) is
    shown in blue. Right: Inference on the inverse problem: Prior with
    $\mu_{\gen} = (1,1)\Trans$, $\sigma_{\gen 1} = \sigma_{\gen 2} =
    1$ and $\rho=-0.5$. Data on $m$ with $\mu_{\mi} = 1$,
    $\sigma_{\mi}=1$ gives the posterior in red. Note the bimodality
    arising in this particular case.}
  \label{fig:illustrations}
\end{figure}

\subsection{Moment Matching}
\label{sec:moment-matching}

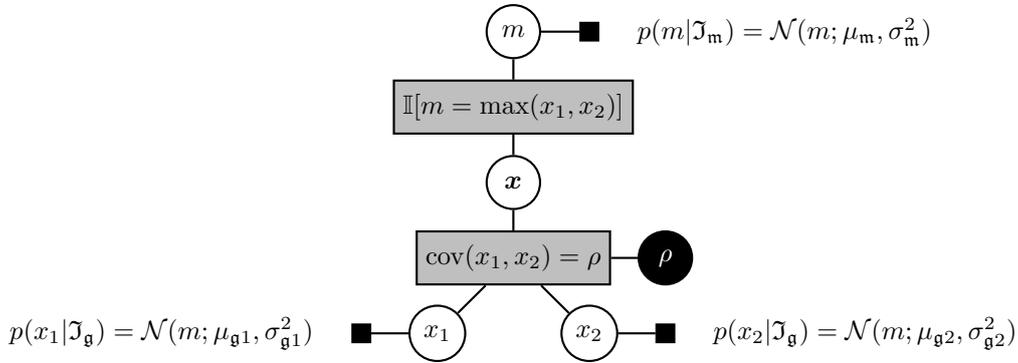
\begin{figure}%
	\begin{center}
	\begin{tikzpicture}[node distance = 1cm, >=stealth',thick]
  \node at (0,0) [var] (m) {$m$};
  \node [prior] (mp) [right of = m] {} edge (m);
  \node [right of=mp,anchor=west,xshift=-.5cm] {$p(m|\I_\mi)=\N(m;\mu_\mi,\sigma^2 _\mi)$};
  \node [factor]    (maxf) [below of=m] {$\Ind[m=\max(x_{1},x_{2})]$}
  edge (m);
  \node [var] (x1x2) [below of=maxf] {$\vec{x}$}
  edge (maxf);
  \node [factor]    (mvf) [below of=x1x2] {$\operatorname{cov}(x_{1},x_{2}) = \rho$}
  edge (x1x2);
  \node [obs] (rho) [right of=mvf, xshift=1cm] {$\rho$} edge (mvf);
  \node [var] (x1) [below of=mvf, xshift=-1cm] {$x_{1}$} edge (mvf);
  \node [prior] (x1p) [left of=x1] {} edge (x1);
  \node [left of=x1p,anchor=east,xshift=.5cm] {$p(x_1|\I_\gen)=\N(m;\mu_{\gen1},\sigma^2 _{\gen1})$};
  \node [var] (x2) [below of=mvf, xshift=1cm] {$x_{2}$} edge (mvf);
  \node [prior] (x2p) [right of=x2] {} edge (x2);
  \node [right of=x2p,anchor=west,xshift=-.5cm] {$p(x_2|\I_\gen)=\N(m;\mu_{\gen2},\sigma^2 _{\gen2})$};
\end{tikzpicture}
\end{center}
\caption{Factor graph representation of the functional relationships
in the inference problems}%
\label{fig:factor-graph_two}%
\end{figure}

The analytical forms derived in the preceeding sections are clearly
not members of the normal exponential family. If $\vec{x}$ has more
than two elements, they also quickly take on complicated forms that
are expensive to evaluate. If the application in question allows,
it might thus be desirable to find Gaussian approximations to the 
posteriors. This section contains derivations for the first
two moments of both posteriors. The Gaussian distributions $q$ matching
these moments minimize the Kullback-Leibler divergence 
$D_{\text{KL}}(p||q) = \int p(y) \log(p(y)/q(y)) \d y$ to the correct
posterior $p$ within the Gaussian family \cite[see, e.g.][Section 10.7]{bishop2006pra}).

\subsubsection{Forward Problem}

We will denote the mean and variance of the posterior of the max as $\mu_{m(12)}$ and $\sigma_{m(12)} ^2$ for reasons that will become clear in Section \ref{sec:max_many}. The corresponding integrals to solve are
\begin{equation}
	\begin{aligned}
	\langle m \rangle &\equiv \mu_{m(12)} = \int_{-\infty} ^\infty m p(m|\I_\com) \d m = Z^{-1} \int m (\nu_1 + \nu_2) \d m\\
	\langle m^2\rangle - \langle m\rangle ^2 & \equiv \sigma_{m(12)} ^2 = \int_{-\infty} ^\infty m^2 p(m|\I_\com) \d m
	\end{aligned}	
\label{eq:18}
\end{equation}
Comparison with Equation \eqref{eq:12} shows that these two integrals are solved by Equation \eqref{eq:tm_gen}. The solutions are thus, after some algebra,
\begin{equation}
\begin{aligned}
\mu_{m(12)} &= w_{1} \left[ \mu_{\com 1}+ \sigma_{\com 1} \frac{b_1}{a_{1}} \frac{\phi(k_{1})}{ \Phi(k_{1}) } \right] + w_{2} \left[ \mu_{\com 2} + \sigma_{\com 2} \frac{b_2}{a_{2}} \frac{\phi(k_{2})}{\Phi(k_{2})} \right]\\
\sigma_{m(12)} ^2 &= w_1 \left\{ \left[\mu_{\com 1} ^2 + \sigma_{\com 1} ^2\right] + \left[ 2\mu_{\com 1} \sigma_{\com1} \frac{b_1}{a_1} - k_1 \sigma_{\com 1} ^2 \frac{b_1 ^2}{a_1 ^2} \right] \frac{\phi(k_1)}{\Phi(k_1)} \right\} \\
      	& \qquad + w_2 \left\{ \left[\mu_{\com 2} ^2 + \sigma_{\com 2} ^2\right] + \left[ 2\mu_{\com 2}\sigma_{\com 2} \frac{b_2}{a_2} - k_2 \sigma_{\com 2} ^2 \frac{b_2 ^2}{a_2 ^2}\right]  \frac{\phi(k_2)}{\Phi(k_2)} \right\} - \mu_{m(12)} ^2
\end{aligned}
\label{eq:19}
\end{equation}
where\marginpar{$w_i,a_i,b_i$}
\begin{xalignat}{2}
\label{eq:20}
      w_{1} &= Z^{-1}\N(\mu_{\mi};\mu_{\gen1},\sigma_{\mi} ^{2} + \sigma_{1} ^{2}) \Phi(k_{1}) &
      w_{2} &= Z^{-1}\N(\mu_{\mi};\mu_{\gen2},\sigma_{\mi} ^{2} +
        \sigma_{2} ^{2}) \Phi(k_{2})\\
			a_{1} &= \left[\sigma_{\gen1}^{2}\sigma_{\gen2}^{2}(1-\rho^{2}) +
        (\sigma_{\gen1}-\rho\sigma_{\gen2})^{2}\sigma_{\com 1} ^{2}
      \right]^{1/2}&
    	a_{2} &= \left[ \sigma_{\gen1}^{2}\sigma_{\gen2}^{2}(1-\rho^{2}) + (\sigma_{\gen2}-\rho\sigma_{\gen1})^{2}\sigma_{\com 2} ^{2} \right]^{1/2}\\
    	b_1 &= \sigma_{\com 1} (\sigma_{\gen1} - \rho \sigma_{\gen2}) & 	
    	b_2 &= \sigma_{\com 2} (\sigma_{\gen2} - \rho \sigma_{\gen1})
\end{xalignat}

\subsubsection{Inverse Problem}

The derivation for the inverse problem is just slightly more involved. We are interested in the moments of the marginals $p(x_1|\I_\com)$ and $p(x_2|\I_\com)$, and will denote these means and variances with $\mu_{1(m2)}$, $\sigma_{1(m2)} ^2$, et cetera. From Equation \eqref{eq:16}, we get
\begin{equation}
\begin{aligned}
\mu_{1(m2)}=\langle x_1 \rangle_{\I_\com} &= \int_{-\infty} ^\infty x_1 \int_{-\infty} ^{x_1}
  \N(x_1;\mu_\mi,\sigma_\mi ^2) \N(\vec{x};\vec{\mu}_\gen,\vec{\Sigma}
  _\gen) \d x_2 \d x_1 \\
  &\quad+ \int_{-\infty} ^\infty \int_{-\infty} ^{x_2}  x_1 \N(x_2;\mu_\mi,\sigma_\mi ^2) \N(\vec{x};\vec{\mu}_\gen,\vec{\Sigma} _\gen) \d x_1 \d x_2
\end{aligned}
\label{eq:21}
\end{equation}
The first integral is in fact identical to the first term of $\mu_{m(12)}$. The second term, however, involves the first \emph{incomplete} moment:
\begin{equation}
\begin{aligned}
	&\int_{-\infty} ^\infty \int_{-\infty} ^{x_2}  x_1 \N(x_2;\mu_\mi,\sigma_\mi ^2) \N(\vec{x};\vec{\mu}_\gen,\vec{\Sigma} _\gen) \d x_1 \d x_2 \\
	&=	 	\int_{-\infty} ^\infty  \N(x_2;\mu_\mi,\sigma_\mi ^2) \N(x_2;\mu_{\gen2},\sigma_{\gen 2} ^2) \int_{-\infty} ^{x_2} x_1 \N\left[x_1; \mu_{\gen 1} + \rho \frac{\sigma_{\gen 1}}{\sigma_{\gen 2}} (x_2 - \mu_{\gen 2}), \sigma^2 _{\gen 1}(1-\rho^2)\right]  \d x_1 \d x_2
\end{aligned}
\label{eq:22}
\end{equation}
The inner integral can be solved using the result given in Equation \eqref{eq:im_gen}, leading to an expression solved by Equation \eqref{eq:tm_gen}. After a bit of algebra, we arrive at the final result
\begin{equation}
    \label{eq:23}
    \begin{aligned}
    \mu_{1(m2)} &= w_{1} \left[\mu_{\com 1}  +  \sigma_{\com 1} \frac{b_1}{a_{1}} \frac{\phi(k_{1})}{ \Phi(k_{1}) } \right] + w_{2} \left[ \left(\mu_{\gen 1} + \rho \frac{\sigma_{\gen 1}}{\sigma_{\gen 2}} (\mu_{\com 2} - \mu_{\gen 2})\right) + \frac{A}{a_{2}}\frac{\phi(k_{2})}{\Phi(k_{2})} \right]\\
    \sigma^2 _{1(m2)} &= w_1 \left\{ \left[\mu_{\com 1} ^2 + \sigma_{\com 1} ^2\right] + \left[ 2\mu_{\com 1} \sigma_{\com1} \frac{b_1}{a_1} - k_1 \sigma_{\com 1} ^2 \frac{b_1 ^2}{a_1 ^2} \right] \frac{\phi(k_1)}{\Phi(k_1)} \right\} \\
      		& \qquad + w_2 \left\{ \sigma_{\gen 1} ^2  \left[ \left(\frac{\mu_{\gen 1}}{\sigma_{\gen 1}} + \rho \frac{(\mu_{\com 2} - \mu_{\gen 2} )}{\sigma_{\gen 2}} \right)^2 + (1 - \rho^2) + \rho^2 \frac{\sigma_{\com 2} ^2 }{\sigma^2 _{\gen 2}} \right] \right.\\
      		& \qquad \left. + \left[\frac{B}{h(1 + \sigma_{\com 2} ^2/h^2)^{1/2}} - \frac{C}{h^3(1 + \sigma_{\com 2} ^2/ h^2)^{3/2}}\right] \frac{\phi(k_2)}{\Phi(k_2)} \right\} - \mu_1 ^2
    \end{aligned}
    \end{equation}
where 
	\begin{equation}
	\begin{aligned}
    \label{eq:24}
    A &= \rho\sigma_{\com 2} ^2 \sigma_{\gen 1} \left(1-\rho \frac{\sigma_{\gen 1}}{\sigma_{\gen 2}}\right) - \sigma_{\gen 1} ^2 \sigma_{\gen 2} (1-\rho^2) \\
    B &= 2\rho^2 \frac{\sigma_{\gen 1} ^2}{\sigma_{\gen 2}^2}\sigma_{\com 2} ^2 (\mu_{\com 2} - \mu_{\gen 2}) + \rho \frac{\sigma_{\gen 1}}{\sigma_{\gen 2}} \left(2\sigma_{\com 2} ^2 \mu_{\gen 1} + \mu_{\gen 2} \frac{\sigma_{\gen 1} ^2 \sigma_{\gen 2} (1-\rho^2)}{\sigma_{\gen 2} - \rho \sigma_{\gen 1}}\right) - \mu_{\gen 1} \sigma_{\gen 1}^2 (1-\rho^2) \frac{\sigma_{\gen 2}}{\sigma_{\gen 2} - \rho \sigma_{\gen 1}}\\
	C &= \rho^2 \frac{\sigma_{\gen 1} ^2}{\sigma_{\gen 2}^2}\sigma_{\com 2} ^4 (\mu_{\com 2} - f) + \sigma_{\gen 1} ^2 (1-\rho^2) \left(1+\rho\frac{\sigma_{\gen 1}}{\sigma_{\gen 2}} \right) \frac{\sigma_{\gen 2}}{\sigma_{\gen 2} - \rho \sigma_{\gen 1}} (\mu_{\com 2} h^2 + f \sigma_{\com 2} ^2)
\end{aligned}
\end{equation}  
with 
\begin{equation}
f = \frac{\sigma_{\gen 2} \mu_{\gen 1} - \rho \sigma_{\gen 1}\mu_{\gen 2}}{\sigma_{\gen 2} - \rho \sigma_{\gen 1}} \qquad \text{and} \qquad
h = \frac{\sigma_{\gen 1} \sigma_{\gen 2} (1-\rho^2)^{1/2}}{\sigma_{\gen 2} - \rho \sigma_{\gen 1}} \\
\label{eq:25}
\end{equation}
The corresponding result for the posterior marginal on $x_2$ can be derived trivially from these results by exchanging the indices $1$ and $2$. Note that, as mentioned above, the first terms of these mixtures are shared with the posterior for $m$. Intuitively, this can be interpreted as follows: For the posterior on $m$, the first term ($\nu_1$) corresponds to the statement that ``if $x_1>x_2$'' (the probability of this is encoded by the cumulative density term in Equation \eqref{eq:12}) ``then $m$ is distributed like $x_1$'' (represented by the product of the probability density functions in \eqref{eq:12}). This part of the relationship features in the inverse problem as well: If $x_1>x_2$, then $x_1$ is distributed like $m$. The second term in the posterior marginal on $x_1$ corresponds to the statement that ``if $x_1<x_2$, then $x_2$ is distributed like $m$ and $x_1$ is distributed such that its distribution fits with the updated marginal of $x_2$ given the correlation between $x_1$ and $x_2$ and the prior marginal on $x_1$.
    
\subsubsection{Related Work}
\label{sec:related_work}
The moments of the likelihood of the max have been derived before by \citet{ClarkMax}. That is, for $\sigma_{\mi} \to \infty$,
the posterior $p(m|\I_\com)$ reported here simplifies to a result reported by Clark:
\begin{equation}
    \label{eq:26}
    \begin{aligned}[t]
      \mu_{m(12)} &\to  \Phi(k) \left[ \mu_{\gen1} +  \sigma_{\gen1} \frac{(\sigma_{\gen1} - \rho\sigma_{\gen2})}{a} \frac{\phi(k)}{\Phi(k)}\right]  + \Phi(-k)\left[\mu_{\gen2} + \sigma_{\gen2}\frac{(\sigma_{\gen2}-\rho\sigma_{\gen1})}{a} \frac{\phi(-k)}{\Phi(-k)}\right] \\
      \sigma_{m(12)} ^2 &\to \Phi(k) \left\{ \left[\mu_{\gen1} ^2 + \sigma_{\gen1} ^2\right] + \left[ 2\mu_{\gen1}\sigma_{\gen1} \frac{(\sigma_{\gen1} - \rho\sigma_{\gen2} )}{a} - k\sigma_{\gen1} ^2 \frac{(\sigma_{\gen1} - \rho \sigma_{\gen2})^2}{a^2} \right] \frac{\phi(k)}{\Phi(k)} \right\} \\
      	& \qquad + \Phi(-k) \left\{ \left[\mu_{\gen2} ^2 + \sigma_{\gen2} ^2\right] + \left[ 2\mu_{\gen2}\sigma_{\gen2} \frac{(\sigma_{\gen2} - \rho\sigma_{\gen1})}{a} + k\sigma_{\gen2} ^2\frac{(\sigma_{\gen2}- \rho \sigma_{\gen1})^2}{a^2}\right]  \frac{\phi(-k)}{\Phi(-k)} \right\} \\ 
      	& \qquad - \mu_{m(12)} ^2 
      	\\ \text{where} \quad
      a &= \sqrt{\sigma^{2} _{\gen1} + \sigma^{2} _{\gen2} - 2\rho \sigma_{\gen1}
        \sigma_{\gen2}} \qquad\text{and}\qquad k = \frac{\mu_{\gen1} -
        \mu_{\gen2}}{a}
    \end{aligned}
\end{equation}
As expected, the posterior of the inverse problem simply becomes equal to the prior in this case. From Equation \eqref{eq:23} we find
\begin{equation}
	\begin{aligned}
	\mu_{1(m2)} &\to \Phi(k) \mu_1 + \sigma_1 \frac{\sigma_1 - \rho \sigma_2}{a} \phi(k) + \Phi(-k) \mu_1 - \sigma_1 \frac{\sigma_1 - \rho \sigma_2}{a} \phi(-k)\\
	&= \Phi(k) \mu_1 + \sigma_1 \frac{\sigma_1 - \rho \sigma_2}{a} \phi(k) + (1-\Phi(k))\mu_1 - \sigma_1 \frac{\sigma_1 - \rho \sigma_2}{a} \phi(k)\\
	&= \mu_1
	\end{aligned}
\label{eq:well-behaved}
\end{equation}
and similarly for the variance. 

The max-factor is also part of the Infer.{\sc net} software package \citep{InferNET} (to my knowledge, the derivations for this code have not been published yet). However, their implementation can only handle two independent Gaussian inputs (Section \ref{sec:max_many} introduces the max over a finite set of correlated variables). So their implementation corresponds to the case of $\rho=0$, which leads to the following simplifications, presented here for reference:
\begin{xalignat}{3}
	k_1 &= \frac{\mu_{\com 1} - \mu_{\gen 2}}{(\sigma_{\gen 1} + \sigma_{\com 2})^{1/2}} &
	a_1 &= \sigma_{\gen 1} (\sigma_{\gen 1} + \sigma_{\com 2})^{1/2} &
	b_1 &= \sigma_{\com 1} \sigma_{\gen 1} \\
	A   &= \sigma_{\gen 1}^2 \sigma_{\gen 2} & 
	B		&= - \mu_{\gen 1} \sigma_{\gen 1} ^2 &
	C 	&= \sigma_{\gen 1} ^2 (\mu_{\com 2} \sigma_{\gen 1}^2 + \mu_{\gen 1}\sigma_{\com 2} ^2) \\
	f   &= \mu_{\gen 1} & h &= \sigma_{\gen 1} & 
\label{eq:27}
\end{xalignat}
Figure \ref{fig:illustrations_approx} shows some of these approximations. The parameter settings used in this figure represent a worst case (e.g., the posterior over $\vec{x}$ is rarely so strongly bimodal.)

\begin{figure}[ht]
  \centering
   \includegraphics[width=0.49\textwidth]{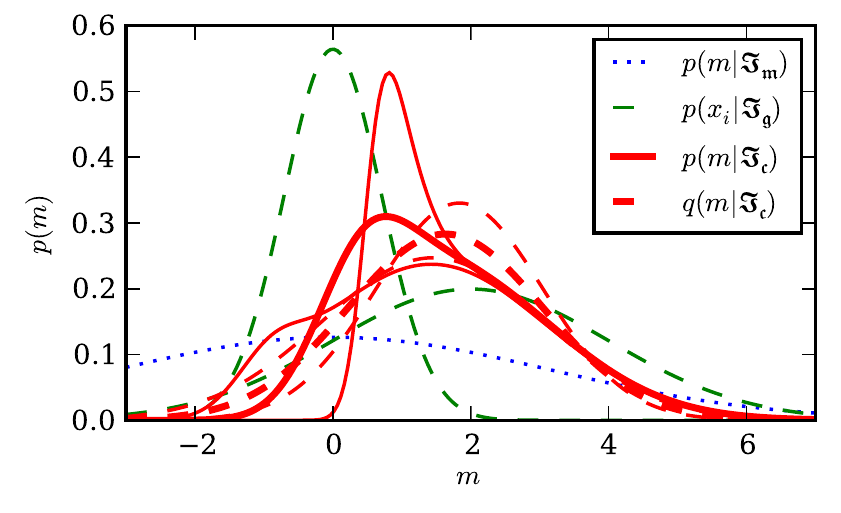}
  \includegraphics[width=0.49\textwidth]{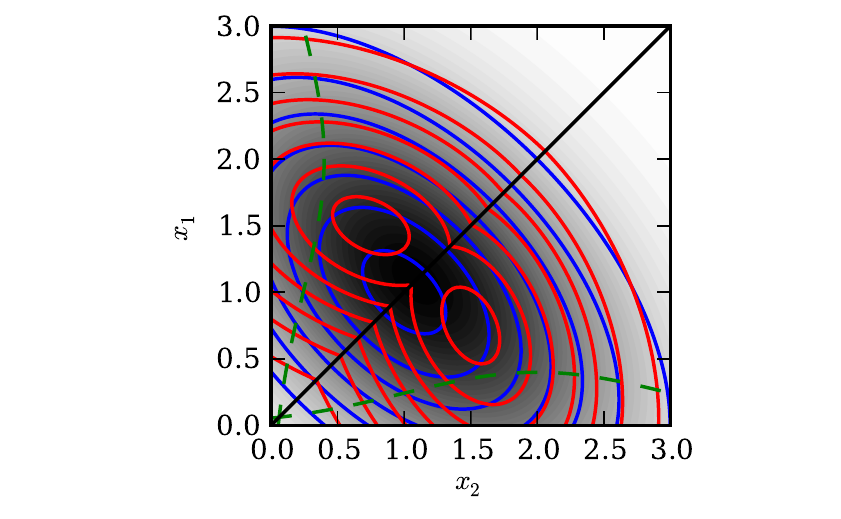}
  \caption{Illustrative plots for the Gaussian approximations to the posteriors. Same beliefs in
  $\I_\com$ as in Figure \ref{fig:illustrations}. Left: For the sake of readability, only the cases $\rho=-0.9$ (broadest),
  $\rho=0$ and $\rho=0.9$ are plotted here. In red, dashed lines the corresponding three Gaussian approximations. Note the
  varying quality of fit. Right: Gaussian approximation (with $\vec{\mu}_{1(m2)}=1.06$ and $\sigma_{1(m2)} ^2 = 0.94$) 
  indicated by shaded area.}
  \label{fig:illustrations_approx}
\end{figure}

\section{The Maximum of a Finite Set}
\label{sec:max_many}

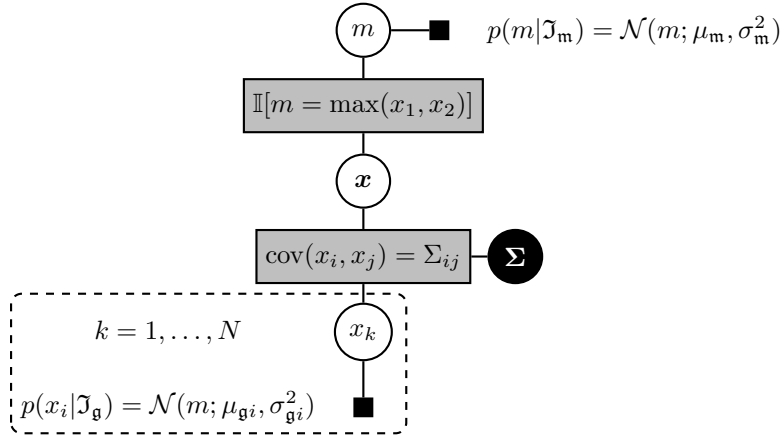
\begin{figure}%
\begin{center}
		\begin{tikzpicture}[node distance = 1cm, >=stealth',thick]
  \node at (0,0) [var] (m) {$m$};
  \node [prior] (mp) [right of = m] {} edge (m);
  \node [right of=mp,anchor=west,xshift=-.5cm] {$p(m|\I_\mi)=\N(m;\mu_\mi,\sigma^2 _\mi)$};
  \node [factor]    (maxf) [below of=m] {$\Ind[m=\max(x_{1},x_{2})]$}
  edge (m);
  \node [var] (x1x2) [below of=maxf] {$\vec{x}$}
  edge (maxf);
  \node [factor]    (mvf) [below of=x1x2] {$\operatorname{cov}(x_{i},x_{j}) = \Sigma_{ij}$}
  edge (x1x2);
  \node [obs] (rho) [right of=mvf, xshift=1cm] {$\vec{\Sigma}$} edge (mvf);
  
  \node[var,below of=mvf] (x) {$x_k$} edge (mvf);
  \node [prior,below of=x] (xp) {} edge (x);
  \node [left of=xp,anchor=east,xshift=.5cm] (xlabel) {$p(x_i|\I_\gen)=\N(m;\mu_{\gen i},\sigma^2 _{\gen i})$};
  \draw[dashed, rounded corners] (xlabel.south west) rectangle +(5.2,1.85);
  \node[above of=xlabel] {$k=1,\dots,N$};
  
\end{tikzpicture}
\end{center}
\caption{Factor graph representation of the inference problem on a finite set. The dashed ``plate'' represents $N$ copies of generating variable nodes.}%
\label{fig:factor_graph_full}%
\end{figure}

\subsection{Analytic Form}
Extending the analysis of Section \ref{sec:Analytic}, we can write the posterior over the max $m$ of a finite set $\{x_i\}_{i=1,\dots,N}$ of variables, distributed according to an $N$-dimensional version of Equation \eqref{eq:1}, with a new normalization constant $Z_N$, as
\begin{equation}
	\begin{aligned}
p(m|\I_\com) &= Z_{N} p(m|\I_\mi)\int p(\vec{x}|m) p(\vec{x}|\I_\gen) \d \vec{x}\\
&= Z_N \N(m;\mu_\mi,\sigma_\mi) \left[\sum_{i=1} ^N \int_{-\infty} ^\infty \delta(m-x_i) p(x_i|\I_\gen) \idotsint_{-\infty} ^{x_i} p(\{x_j\}_{j\neq i} | x_i,\I_\gen) \prod_{j\neq i}\d x_j \d x_i  \right]\\
&=Z_N \sum_{i} \left[\N(\mu_\mi;\mu_{\gen i},\sigma_\mi ^2 + \sigma_{\gen i} ^2) \N(m;\mu_{\com i},\sigma_{\com i} ^2) \idotsint_{-\infty} ^{x_i} 
	\N(\vec{x}_{\wo i}; \vec{\mu}_{\gen \wo i}(x_i),\vec{\Sigma}_{\gen \wo i}) \d \vec{x}_{\wo i} \right]
	\end{aligned}
\label{eq:28}
\end{equation}
where $\vec{x}_{\wo i}=(x_1,\dots,x_{i-1},x_{i+1},\dots,x_N)$. The conditional mean is \citep[see e.g.][Section 2.3.2]{bishop2006pra}
\begin{equation}
	\left(\vec{\mu}_{\wo i}(x_i)\right)_j = \mu_{\gen j} + \Sigma_{\gen ji} \Sigma_{\gen ii} ^{-1} (x_i - \mu_{\gen i}) = \mu_{\gen j} + \rho_{ij} \frac{\sigma_{\gen j}}{\sigma_{\gen i}} (x_i - \mu_{\gen i})
\label{eq:29}
\end{equation}
with the linear coefficient of correlation $\rho_{ij} = \Sigma_{\gen ij} / (\sigma_{\gen i}\sigma_{\gen j})$. The conditional covariance matrix is the Schur complement of $\Sigma_{\gen ii}=\sigma_{\gen i} ^2$ in $\vec{\Sigma_{\gen}}$:
\begin{equation}
\Sigma_{\gen \wo i, kj} = \Sigma_{\gen kj} - \Sigma_{\gen ki} \sigma_{\gen i} ^{-2} \Sigma_{\gen ij}
\label{eq:30}
\end{equation}
In principle, it would be possible to follow the path laid out in the previous sections to calculate the first two moments of this distribution. However, while the univariate Gaussian CDF (essentially an evaluation of the error function) has computational cost comparable to evaluating an exponential function, computationally efficient ways of calculating a multivariate Gaussian CDF are not generally available. So this approximation would need to involve an undesirable numerical integration.

\subsection{A Heuristic Approximation}

Another, cheaper option is to use an iterative procedure initially proposed by \citet{ClarkMax}. The idea is to start out with the approximation for only two of the generating variables. W.l.o.g., let these be $x_1$ and $x_2$, resulting in $m_{(12)}=\max(x_1,x_2)$. Next, estimate $m_{(123)}=\max(x_3,m_{(12)})$ and so on up to $m_{(1\dots N)}$. For the intermediate maxima, the likelihoods presented in Equation \eqref{eq:26} suffice, and the prior is included in the last step (using Equation \eqref{eq:19}) to gain an approximate posterior over the maximum of the whole set. Of course, this necessitates an analytic expression for the correlation coefficient $\rho_{i(1\dots i-1)}$ between the $i$-th variable and the max over the preceding variables. This was derived by Clark. Adopted to the notation used here and made more explicit, his result is
\begin{equation}
	\rho_{3(12)} = \sigma_{(12)} ^{-1} \left( \sigma_{1} \rho_{31} \Phi(k_{(12)}) + \sigma_2 \rho_{32} \Phi(-k_{(12)})\right)
\label{eq:31}
\end{equation}
where $\rho_{ij} = \Sigma_{ij} / \sigma_i \sigma_j$, the index $\gen$ has been dropped for simplicity and $k_{(12)}=(\mu_1 - \mu_2)/\sqrt{\sigma_1 ^2 + \sigma_2 ^2}$ is the simplified version of $k_{1}$ arising from Equation \eqref{eq:k_norm} under $\sigma_\mi \to \infty$. Using this, we can build a recursive algorithm to calculate $\rho_{i(1\dots j)}$ with $j<i$ as
\begin{equation}
\rho_{i(1\dots j)} = 
	\sigma_{(1\dots j)} ^{-1} \cdot \begin{cases} 
		\sigma_j \rho_{ij} & \text{if~}j = 1 \\
	  \Phi(-k_{(1\dots j)}) \sigma_{j} \rho_{ij} + \Phi(k_{(1\dots j)}) \rho_{i(1\dots j-1)}  & \text{else}
	\end{cases}
\label{eq:32}
\end{equation}
this necessitates a list $\vec k_{[j]} = (k_{(12)}, \dots k_{(1\dots i-1)})$ which is available at the necessary point in time from the calculation of previous maxima over the preceding parts of the set. Note that calculating $\rho_{i(1\dots i-1)}$ involves $i-1$ recursive function calls, so building the full approximation over the max of $N$ variables is of complexity $\mathcal{O}(N^2)$, as might be expected (although there are only $(N-1)$ uses of the results in Equation \eqref{eq:19}). If all correlation coefficients are the same, $\rho_{ij} = \rho\:\forall ij$, then the recursive evaluations can be re-used in consecutive evaluations and the complexity drops to $\mathcal{O}(N)$.

\subsubsection{Inverse Problem}

The same iterative scheme can be used to provide an approximation for the inverse problem's posterior. First, the list $\vec{k}_{[j]}$ is build as in the preceding section. Then, approximations to the posterior marginals are build iteratively, starting with $q(x_N|\I_\com)$, ending with $q(x_2|\I_\com)$ and $q(x_1|\I_\com)$. At each intermediate step, we use the EP approximation \citep{minka2001epa}: To get $q(x_i|\I_\com)$, use $q(m_{(1\dots i)}|\I_\mi)=q(m_{(1\dots i)}|\I_\com)/q(m_{(1\dots i)}|\I_\gen)$ as an approximation to the prior over the subset max, and $q(m_{(1\dots i-1)}|\I_\gen)$ as the approximation on the max over the subset up to $x_{i-1}$.

\section{Discussion of the Approximation's Quality}
\label{sec:Discussion}

\begin{figure}%
\includegraphics[width=\textwidth]{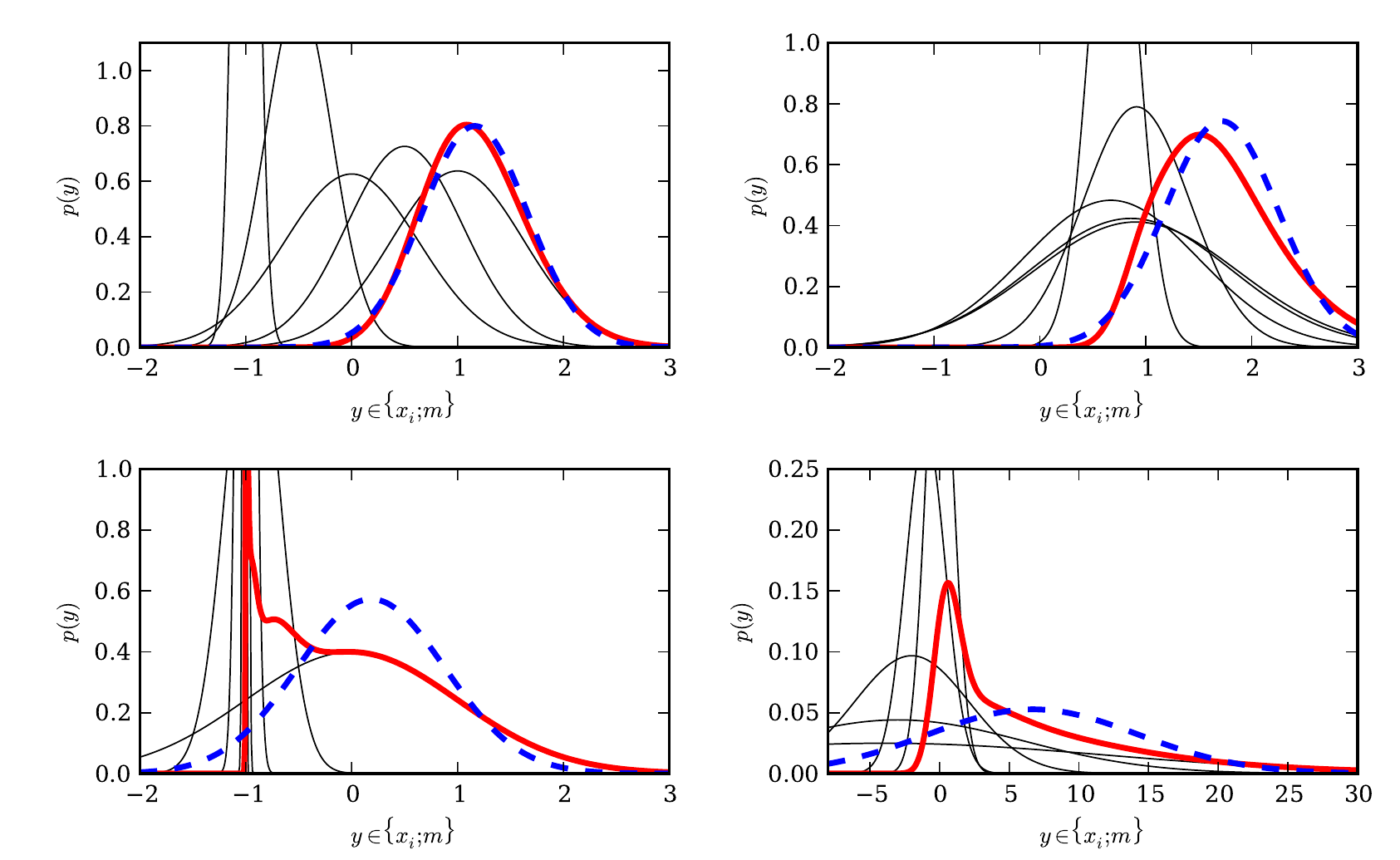}%
\caption{Quality and failure modes of the EP approximation. Max of five uncorrelated Gaussians. Top row: examples of good fits. Left: well separated beliefs. Right: similar beliefs. Bottom row: worst case examples. Left: high certainty contributions within the center. Right: high uncertainty in one tail. In all plots, beliefs over the $x_i$ as slim black lines. True posterior over $m$ in thick red, approximation in thick dashed blue. For simplicity, $p(m|\I_\mi)$ was set to an uninformative value. See text for details.}%
\label{fig:approx_q}%
\end{figure}

Figure \ref{fig:approx_q} gives some intuition on the quality of the approximation. For the purpose of this comparison, uncorrelated Gaussians were used because this allows the analytic evaluation of the true posterior (the CDF factorises into individual one-dimensional CDFs). The fit is reasonably good if the beliefs over the $x_i$ are either very similar (Figure \ref{fig:approx_q} top right), or if the beliefs are ``separated'', in the sense that one of the $x_i$ provides a dominant contribution to the overall mixture (top left). The fit becomes bad when the mixture has many modes (bottom left) or a strong asymmetry (bottom right). The corresponding worst case distributions shown here were generated by setting $\mu_{\gen i} = a + b^{-i}$ and $\sigma_{\gen i} ^2 = b^{-i}$ (left, $a=-1, b=16$) or $\mu_{\gen i} = ci$ and $\sigma_{\gen i} ^2 = i^d + 1$ (right, $c=-1,d=16$). More quantitatively, consider Equation \eqref{eq:26} or Equation \eqref{eq:12}, the case of the max of only two Gaussians. The two cases of good fit described above correspond to
\begin{enumerate}
	\item one mixture component dominating the mixture
	\begin{equation}
	|k_{12}| = \frac{| \mu_{\gen 1} - \mu_{\gen 2}|}{\sqrt{\sigma_{\gen 1} ^2 + \sigma_{\gen 2} ^2 - 2 \rho \sigma_{\gen 1} \sigma_{\gen 2}}}\gg 0
\label{eq:33}
\end{equation}
	The likelihood then has one clearly dominating Gaussian component and the fit is good. In this case, the inverse problem is also a good fit, as each of the generating variables $x_1,x_2$ has one dominating component in its posterior.
  \item the two mixture components being almost identical:
 	\begin{equation}
		\mu_{\gen 1} \approx \mu_{\gen 2} \qquad \text{and} \qquad \sigma_{\gen 1} \approx \sigma_{\gen 2}
		\label{eq:34}
	\end{equation}
	the likelihood then consists of two roughly identical Gaussian components with roughly the same weights, and is therefore roughly Gaussian. However, the approximation is \emph{bad} for the inverse problem here, as the true posterior marginals become bimodal (c.f. Figure \ref{fig:illustrations}, right). This effect is particularly pronounced if the mean of the prior and the likelihood differ significantly.
\end{enumerate}
These observations suggest a potential increase in the quality of the approximation to be gained from calculating all $N(N-1)$ weight-generators $k_{ij}$ as defined in Equation \eqref{eq:33} and iteratively choosing the pair $ij$ with maximal $k_{ij}$. However, this re-ordering has to be updated after each incremental two-component max operation, involving a re-calculation of up to $N$ correlation coefficients. It thus raises the complexity of calculating the approximation for the overall max from $\mathcal{O}(N^2)$ to $\mathcal{O}(N^3)$. Initial experiments suggest that the potential gain in fit is almost always negligible.

\begin{figure}[ht]%
\includegraphics[width=\textwidth]{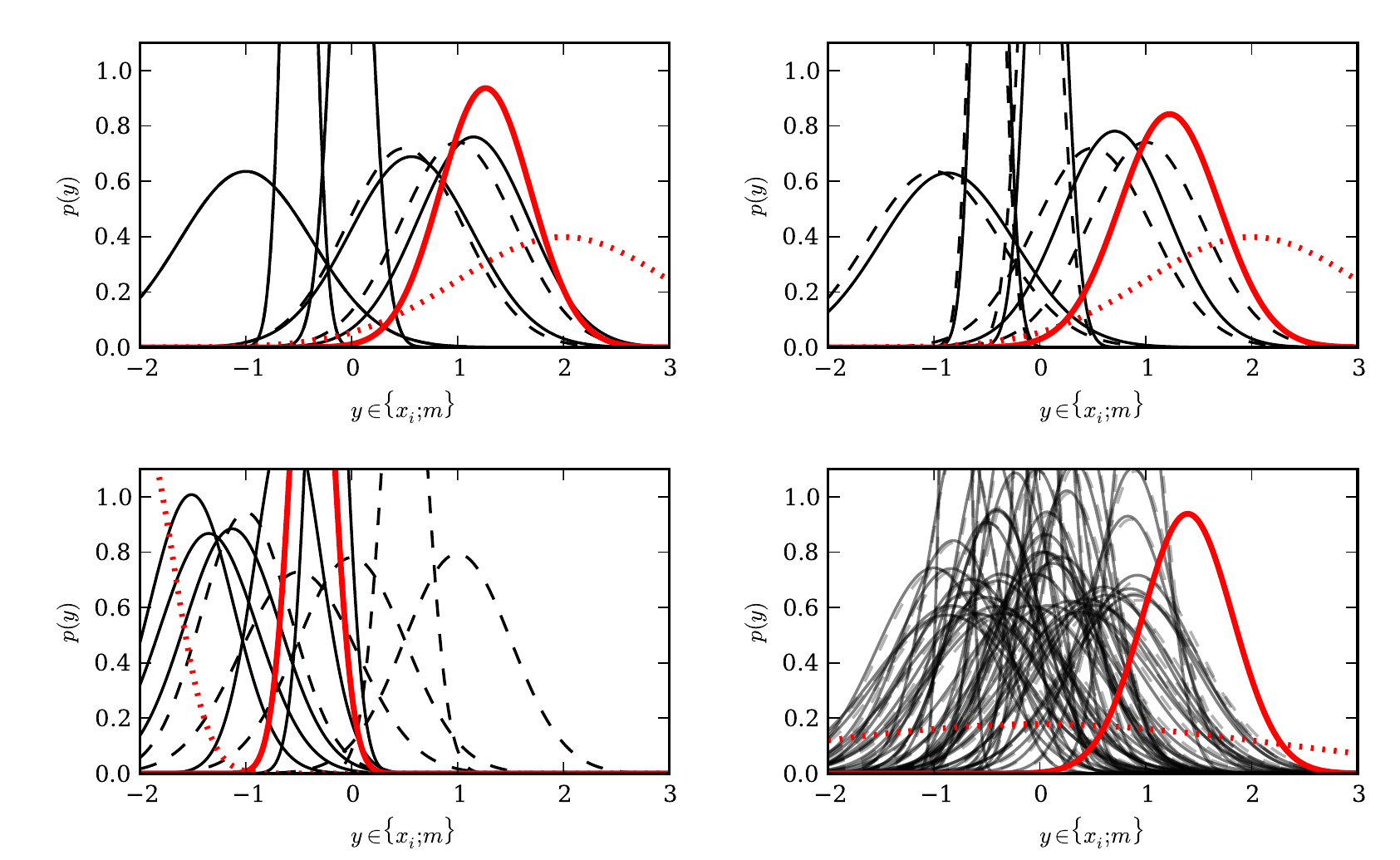}%
\caption{Illustrative examples for the use of the approximation in EP message passing. $p(x_i|\I_\gen)$ in black dashed lines. $p(m|\I_\mi)$ in red dotted. Marginals after EP message passing as corresponding solid lines. Top left: Max over 5 uncorrelated variables. Only the two variables contributing significantly to the max change their beliefs. Top right: same as previous, but with $\rho_{ij}=0.9$ for all $ij$. The change in belief over the dominating $x_i$ now also effects the other beliefs, as expected. Bottom left: The approximation is well-behaved under inconsistent beliefs. $p(m|\I_\mi)$ was set inconsistently low relative to the beliefs on the $x_i$ (all $\rho_{ij}=0.2$). Note that the belief over the largest $x_i$ extends beyond the belief over $m$ as a result of the moment-matching. Bottom right: The approximation is stable for large values of $N$. Maximum over 50 correlated normals, all $\rho_{ij}$ were set to 0.5.}%
\label{fig:demos}%
\end{figure}


\section{Conclusion}
This technical report derived the first two moments of the posterior over the maximum of a pair of Gaussian variables, and over the posterior over the two generating variables. These moments can be used for approximate Inference on their own, or as part of a larger graphical model using Expectation Propagation. I have also shown how to extend the usefulness of these approximations to finite sets of Gaussian variables using a heuristic iterative approximation. The quality of the approximation depends on the location and precision of the belief over the generating variables relative to each other, but is always good enough to provide a meaningful point estimate and error measure. It is sufficiently robust to deal with inconsistent belief assignments and large numbers of generating variables (see Figure \ref{fig:demos}).

\bibliographystyle{plainnat} \bibliography{bibfile}

\end{document}